\documentclass[letterpaper, 10 pt, conference]{ieeeconf}  % Comment this line out if you need a4paper

\IEEEoverridecommandlockouts                              % This command is only needed if 
\usepackage{graphicx} % Required for inserting images
\usepackage{pgfplots}
\usepackage{standalone}
\usepackage{siunitx}
\usepackage{hyperref}
\usepackage{multirow}
\usepackage[font=small,skip=0pt]{caption}

\newcommand{\omnidirectional}{omnidirectional} % Rember to change the title if this changes.
\newcommand{\fov}{FoV}
\newcommand{\etal}{\textit{et.~al.}}

\title{\LARGE \bf
Geometry-Informed Distance Candidate Selection for Adaptive Lightweight Omnidirectional Stereo Vision with Fisheye Images
}

\author{Conner Pulling$^{1}$, Je Hon Tan$^{2}$, Yaoyu Hu$^{1}$, Sebastian Scherer$^{1}$% <-this % stops a space
\thanks{*This work was supported by the Defence Science and Technology Agency, Singapore.}% <-this % stops a space
\thanks{$^{1}$C. Pulling, Y. Hu and S. Scherer are with the Robotics Institute, Carnegie Mellon University, 5000 Forbes Avenue
        Pittsburgh, PA 15213-3890 USA. {\tt\small \{cpulling, yaoyuh, basti\}@andrew.cmu.edu}.}%
\thanks{$^{2}$J. Tan is with the Defence Science and Technology Agency, 1 Depot Road, Singapore, 109679.
        {\tt\small jehontan@gmail.com}.}%
}

\begin{document}

\maketitle
\thispagestyle{empty}
\pagestyle{empty}

%%%%%%%%%%%%%%%%%%%%%%%%%%%%%%%%%%%%%%%%%%%%%%%%%%%%%%%%%%%%%%%%%%%%%%%%%%%%%%%%
\begin{abstract}

Multi-view stereo \omnidirectional{} distance estimation usually needs to build a cost volume with many hypothetical distance candidates. The cost volume building process is often computationally heavy considering the limited resources a mobile robot has. We propose a new geometry-informed way of distance candidates selection method which enables the use of a very small number of candidates and reduces the computational cost. We demonstrate the use of the geometry-informed candidates in a set of model variants. We find that by adjusting the candidates during robot deployment, our geometry-informed distance candidates also improve a pre-trained model's accuracy if the extrinsics or the number of cameras changes. Without any re-training or fine-tuning, our models outperform models trained with evenly distributed distance candidates. Models are also released as hardware-accelerated versions with a new dedicated large-scale dataset. The project page, code, and dataset can be found at \href{https://theairlab.org/gicandidates/}{https://theairlab.org/gicandidates/}.

\end{abstract}

%%%%%%%%%%%%%%%%%%%%%%%%%%%%%%%%%%%%%%%%%%%%%%%%%%%%%%%%%%%%%%%%%%%%%%%%%%%%%%%%
\section{INTRODUCTION}

Distance perception is a key requirement in mobile robots. A larger field-of-view (\fov{}) and faster distance perception enable a robot to more effectively gather information about its surroundings, with \omnidirectional{} sensing being the most desirable. Presently, LiDAR devices are the go-to sensors for (horizontally) \omnidirectional{} sparse distance perception. It is technically difficult and prohibitively expensive to achieve true \omnidirectional{} \fov{} and high resolution with LiDARs.

Using multiple cameras as a multi-view stereo (MVS) camera set can provide high-resolution \omnidirectional{} distance perception with much lower mechanical complexity and cost. Recent research has demonstrated that using multiple cameras with large \fov{} lenses (e.g., fisheye lens) can achieve \omnidirectional{} distance estimation \cite{won2019omnimvs, meuleman2021real, xie2023omnividar}. Compared to LiDAR devices, vision-based distance estimation typically provides larger \fov{} and denser measurements. However, two challenges prevent MVS-\omnidirectional{} solutions from being the go-to choice: 1) they are computationally expensive and 2) difficult to deploy.

The majority of the MVS-\omnidirectional{} models, both learning-based and non-learning, utilize a cost volume structure that aggregates visual features by using virtual distance candidates along a viewing direction. The model compares the features in the cost volume and picks the best weights for a linear combination of the given candidates. This approach consumes a significant amount of computing resources, which grows depending on the number of cameras and the number of distance candidates.

\begin{figure}[t]
    \centering
    \includegraphics[width=0.9\linewidth]{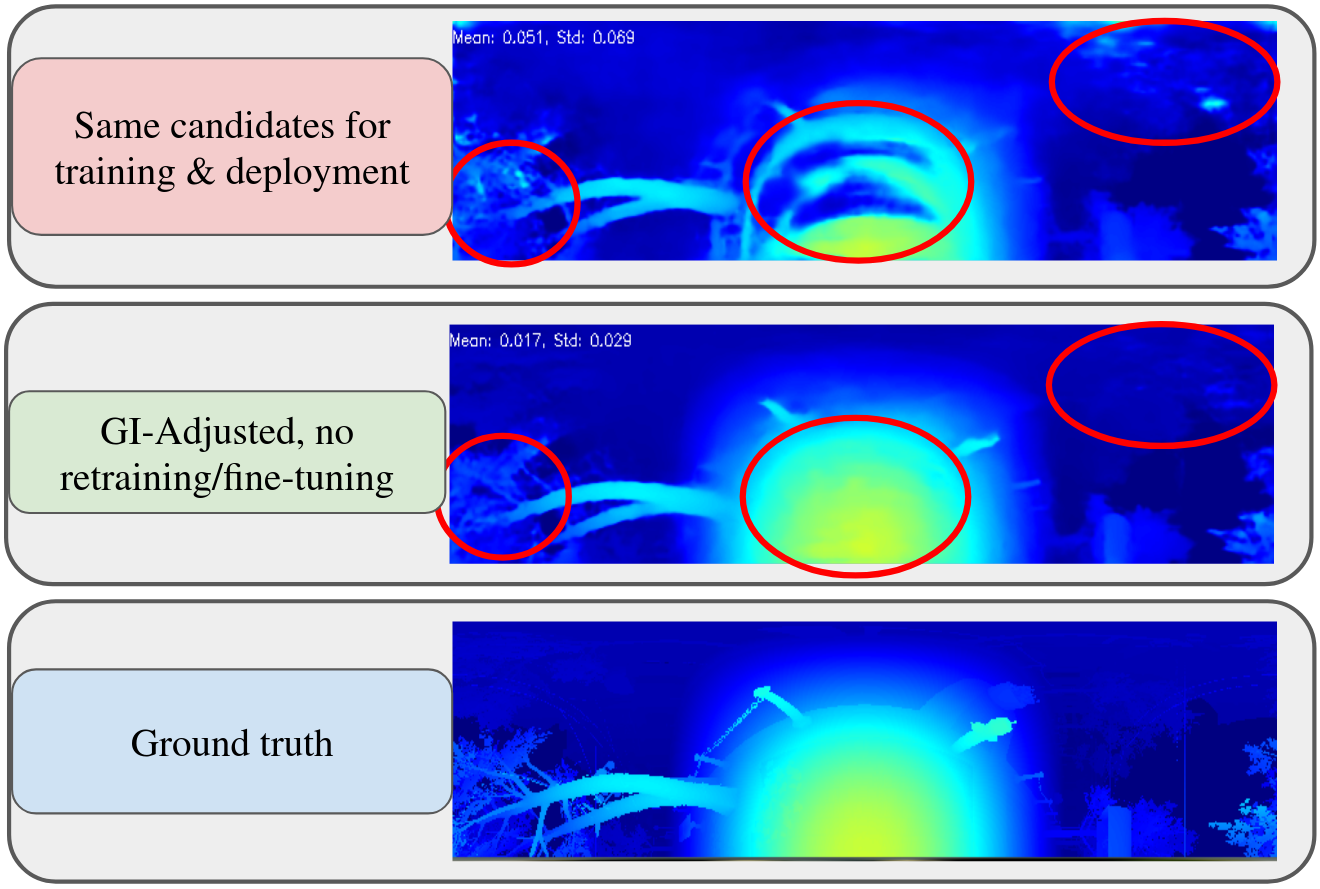}
    \caption{After training, using our geometry-informed (GI) distance candidate distribution, the baseline distance between cameras can be changed and the model's performance can be restored without fine-tuning.}
    \label{fig:abstractbanner}
\end{figure}

For deployment, cameras in an MVS-\omnidirectional{} system typically need to be placed such that maximum \fov{} can be achieved with minimum occlusions from the robot (self-occlusion). For learning-based methods, if the location or number of cameras is changed to mitigate occlusions, the method typically suffers significant performance degradation as the position of corresponding features in the camera images is changed, hence for the same distance candidates the patterns of accumulated features in the cost volume differ greatly from the training data.

To resolve the above issues related to learning-based visual \omnidirectional{} distance estimation, our insight is that we can train a model to utilize a small number of virtual distance candidates by picking distance candidates in a way that is informed by the geometry of the camera configuration. For a known set of camera extrinsics, we can select the candidates such that the positional displacement for the feature sampled at two consecutive distance candidates are similar across all consecutive pairs of candidates, allowing the model to more effectively determine the best interpolation weights between a pair of consecutive candidates. This enables us to create models with a much lower number of candidates (16 or 8) compared to previous methods, significantly reducing computational cost. We are also able to compute such distance candidates for new camera configurations during deployment, allowing a trained model to be used and maintain its performance even if the camera extrinsics or the number of cameras is changed. In this work, our contributions are

\begin{itemize}
    \item A geometry-informed (GI) distance candidates selection method that enables the use of fewer candidates and change of extrinsics for deployment.
    \item Demonstration of a set of GI candidates-trained models on several plane-camera layouts with self-occlusion and variable translations among the cameras.
\end{itemize}

After training on our dedicated new dataset, our model can efficiently generate \omnidirectional{} distance from multiple cameras with self-occlusion explicitly handled, even if the number and position of the cameras change during physical deployment. The code, pre-trained model, and the dedicated dataset are available through the project webpage. 

\section{RELATED WORKS}
Estimating distance from more than one camera is a common and fundamental capability of robot systems. There is a vast body of work that covers various topics, of which we will concentrate on two most closely related to ours.

\subsection{Multi-view Omnidirectional Distance Estimation}
The most relevant non-learning model is from Meuleman, \etal{} \cite{meuleman2021real} where they generate distance predictions for a reference fisheye image by selectively fusing information from other fisheye image views. A complete \omnidirectional{} distance prediction is then made by stitching multiple estimations together. They also build a cost volume to aggregate information across different distance candidates. For efficiency, the number of candidates is kept at 32. Since the model is non-learning-based, there is no training and it can be deployed on various camera layouts. This model is one of our main baseline models.

For the learning-based models, SweepNet and OmniMVS, by Won, \etal{} \cite{won2019omnimvs}\cite{won2019sweepnet}\cite{won2020end} are the standouts among the early approaches. Like the non-learning models, SweepNet and OmniMVS will build a cost volume for a fixed number of candidates. This number is configurable but in order to achieve desired accuracy the value is set at around 100 or 200. The cost volume is consumed by the downstream part of the model, typically layers of 3D Convolutional Neural Networks (CNN), and distance values are estimated. Later, Su, \etal{} \cite{su2023omnidirectional} implemented a hierarchical version that makes distance predictions on different scales, where at each scale, a cost volume is built in the same way. The above models are trained with a fixed number of cameras and placement. When the camera layout changes, new training and datasets may be required.

Two recent works are closely related to our approach. One conducted by Chen, \etal{} \cite{chen2023unsupervised} constructs multiple cost volumes for unsupervised learning. They use feature variance to compare the cost volumes \cite{yao2018mvsnet}. Our approach is similar with the difference being that we handle self-occlusion explicitly. The other is OmniVidar \cite{xie2023omnividar}, which turns \omnidirectional{} distance estimation into multiple rounds of binocular stereo estimations. On a high level, the learning-based part of this approach is camera layout agnostic as long as we can cover the final \omnidirectional{} \fov{} by undistorting and rectifying the input fisheye images along different orientations. However, this process needs to be manually and carefully designed for every new camera layout. Our model can accommodate camera layout change through an easier process with fewer manual procedures.

% SweepNet \cite{won2019sweepnet} also first warps the input fisheye image. In our model, we also do similar pre-processing to focus more on the boundary region of a fisheye image.

\subsection{Multi-view Stereo (MVS)}

Multi-view stereo (MVS) has a longer history compared with the aforementioned multi-view \omnidirectional{} distance estimation. MVS studies are more focused on reconstructing the 3D geometry of an object or a scene, other than providing distance estimations with respect to a robot. Similar to \omnidirectional{} distance estimation, MVS studies use both non-learning \cite{kutulakos2000theory}\cite{seitz1999photorealistic}\cite{furukawa2009accurate}\cite{lhuillier2005quasi}\cite{campbell2008using}\cite{xu2019multi} and learning-based approaches \cite{yao2018mvsnet}\cite{kar2017learning}\cite{yao2019recurrent}\cite{yi2020pyramid}\cite{peng2022rethinking}. The result of an MVS method is usually a volumetric representation (e.g., voxel grid surface), point cloud, or surface mesh. Inside these learning-based models, a cost volume can be constructed following \cite{yao2018mvsnet}. Most of the approaches use a reasonably large number of distance candidates. Some works, e.g. \cite{gu2020cascade}\cite{li2022ds}\cite{gao2022cost}, explore multi-scale or adaptive candidates, which may use fewer candidates but need to do the computing in an iterative way, leading to additional computational overhead.

% MVS models are naturally extrinsics agnostic, meaning the model should be able to handle free camera spacing. If trained with a variable number of camera views, an MVS model can be generalized to change of camera quantity. However, most of the MVS models consider occlusion introduced by the scene and the movement of a robot. For our work, we explicitly held self-occlusion induced by the body of a robot.

\begingroup
\setlength{\belowcaptionskip}{-10pt}

\begin{figure*}[ht!]
    \centering
    \includegraphics[width=0.85\linewidth]{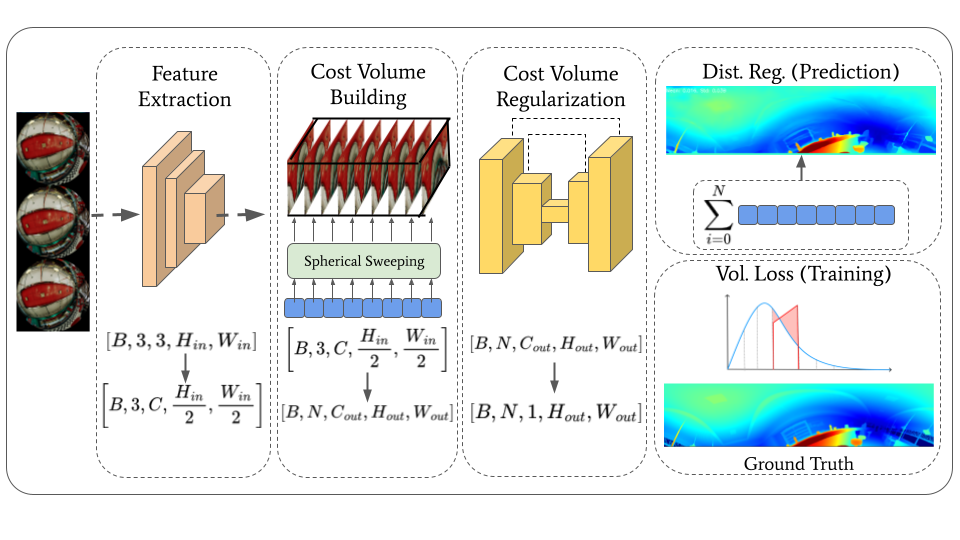}
    \caption{Model Overview. The model takes three fisheye images as input during training and performs learned feature extraction with a shared feature extractor, builds a cost volume with spherical sweeping, and regularizes the distance with a 3D U-Net \cite{won2019omnimvs}.}
    \label{fig:architecture}
\end{figure*}

\endgroup

\section{METHODS}

%\subsection{Justify why we need all cameras facing the same direction}

\begin{figure}[b!]
    \centering
    \includegraphics[width=0.9\linewidth]{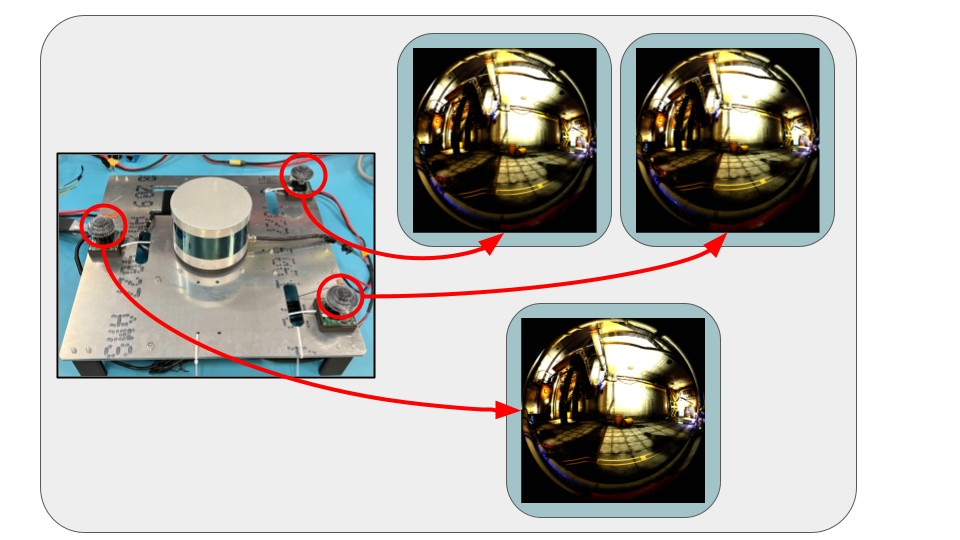}
    \caption{Camera Configuration for the evaluation board. Three fisheye cameras are mounted pointing upwards in a triangular formation. A LiDAR, unused for this study, introduces self-occlusions.}
    \label{fig:hardware}
\end{figure}

\subsection{Target Configuration}
For real-world testing, we use an evaluation board with three fisheye cameras pointed in the same direction and arranged in a triangular formation as in Fig.~\ref{fig:hardware} and the \textit{training layout} in Fig.~\ref{fig:layouts}. This target configuration enables an aerial robot to have \omnidirectional{} vision by placing cameras safely on top of its body, e.g. Skydio 2+ Drone. 
% Additionally, this target configuration is especially challenging for distance estimation due to the high \fov{} overlap, meaning many areas are very visually similar in all camera images.
Additionally, this target configuration is especially challenging due to the fact that image boundary regions from fisheye lenses are extensively used where good calibration is hard to achieve. We utilize the TartanCalib toolbox to get better calibration results with the Double Sphere camera model\cite{duisterhof2022tartancalib}\cite{usenko2020doubsphere}.

\subsection{Model Overview}
Similar to \cite{won2019omnimvs}, our model builds a cost volume from spherically-sweeping learned features and then regularizes this cost volume to achieve a probability distribution of the true distance for each pixel.  First, the model takes in three fisheye images during training. Feature maps are extracted from the images with a shared 2D-convolution feature extractor. Next, spherical sweeping is employed using a set of distance candidates to warp the other fisheye images into the reference image frame at the candidate distance. To aggregate all of the views into $C$ channels, differing from prior works in \omnidirectional{} vision with fisheye images, one of our model variants (introduced in Section \ref{sec:model_variants}) uses feature variance to build the cost volume, similar to \cite{yao2018mvsnet}. By using feature variance as opposed to concatenating the feature vectors together for each pixel for each warped image, the channel dimension is reduced by a factor of $N$ (number of images). Additionally, because the variance between a set of vectors results in a same-length vector no matter how many vectors there are from the input images, the model can explicitly exclude self-occluded pixels while maintaining the required length of the $C$ dimension. 
% By explicitly excluding self-occluded pixels, the regularizer never sees invalid pixels and the learning problem becomes easier. 
Since the cost volume has one dimension more than the shape of the extracted 2D features, operations like 3D convolutions need to be applied. We utilize a 3D U-Net typed regularizer to process the cost volume into a probability distribution. The probability for each candidate is used in a weighted sum to regress the distance for each pixel.

%\subsection{Standard Deviation Cost Volume Aggregation}

%The use of equirectangular projection results in a region of high distortion at the top of the projected image. While conventional planar convolution kernels may effectively extract low-level features (e.g. edges, corners), they will face difficulty with high-level features (e.g. textures, object features) which appear distorted differently in different regions of the image. To account for the projection distortion we use a spherical convolution similar to that described in \cite{Li_Jin2022MODE}, implemented using the deformable convolution with fixed precomputed kernel offsets. Our spherical feature extraction module comprises two ResNet stages of 5 and 10 layers respectively, followed by a spherical convolution to aggregate high-level features. We show that the use of spherical convolution improves distance prediction quality in the high distortion region of the image.

\subsection{Distance candidate selection}

\begin{figure}[!b]
    \centering
    \includegraphics[width=\linewidth]{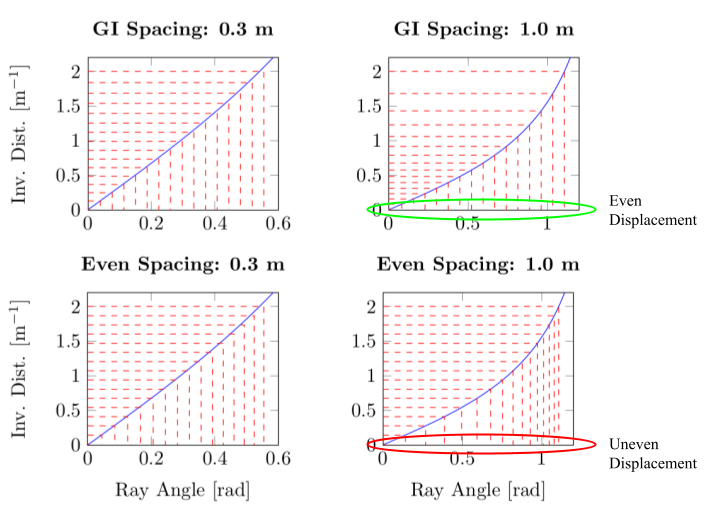}
    \caption{GI and EV candidates for different camera spacings. EV candidates approximate constant feature displacement steps for small spacings (baselines), but result in highly uneven steps in large spacings. GI candidates generate constant displacement steps as a function of camera spacing.}
    \label{fig:candidates}
\end{figure}

\begin{figure}[!t]
    \centering
    \includegraphics[width=0.75\linewidth]{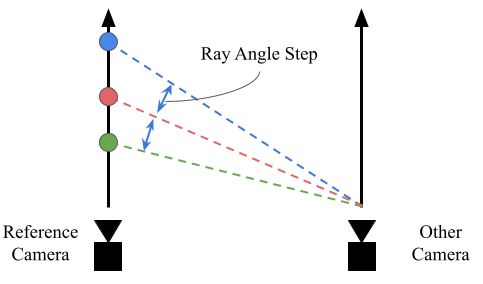}
    \caption{Sphere-sweeping geometry. We pick distance candidates that result in constant steps in ray angle corresponding to constant displacements in the projected feature.}
    \label{fig:cand_geometry}
\end{figure}

Previous work on distance perception commonly used distance candidates spaced evenly in the inverse distance space (hereinafter named \textit{EV}). In the case of plane-sweeping\cite{yao2018mvsnet}, EV candidates have the property that moving an object between consecutive candidates results in a constant pixel displacement of the corresponding features in feature space. 
% This property makes training easier for convolution-based feature-matching methods.

In the case of sphere-sweeping\cite{meuleman2021real}\cite{won2020end}, EV candidates generally do not result in constant feature displacement due to the non-linearity of spherical sweeping. However, for small camera baselines,
%and equirectangular projection
they provide a close approximation, as shown in Fig.~\hyperref[fig:candidates]{\ref*{fig:candidates}}. As previous work on sphere-sweeping has focused on small baseline configurations and large numbers of candidates\cite{won2019omnimvs}\cite{won2019sweepnet}\cite{won2020end}, the use of EV candidates caused negligible impact on performance.

For better efficiency, we propose to use a small number of geometry-informed (GI) candidates computed for specific camera extrinsics and ensure similar displacement for each step between distance candidates. As feature position in the projected image is proportional to the feature ray angle, GI candidates are obtained by developing distance as a function of ray angle and sampling it with evenly spaced ray angle steps (see Fig.~\hyperref[fig:candidates]{\ref*{fig:cand_geometry}}). Later in the experiment section, we show that the use of GI candidates improves distance prediction accuracy in the cases of large camera spacing and low candidate count.

\begin{figure}[!b]
    \centering
    \includegraphics[width=\linewidth]{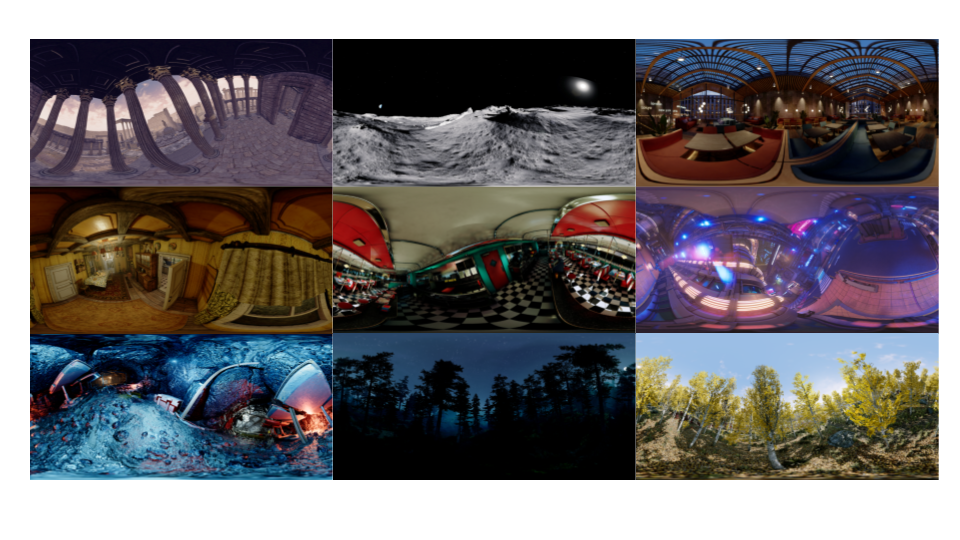}
    \caption{Our omni-directional stereo vision with fisheye images dataset consists of about 95K samples from over 60 Unreal Engine 4 high-fidelity simulation environments, manifested in various scene styles. }
    \label{fig:datasetbanner}
\end{figure}

\subsection{Volume Loss}
As seen in previous work \cite{won2019omnimvs}\cite{xie2023omnividar}\cite{su2023omnidirectional}, the main loss function of choice for the omnidirectional stereo vision supervised learning problem has been L1 loss on the final distance map. However, there is a rich amount of information in the cost volume itself before aggregation. Before linear combination but after softmaxing, the cost volume represents a probability distribution of which distance candidate is the most likely to be the true distance. In actuality, this probability distribution should look like the interpolation between the two closest distance candidates to the true distance value. Therefore, because the ground truth probability distribution is known and the softmax'd cost volume represents a predicted probability distribution, a soft cross-entropy loss function can be used as a more informative loss function \cite{nuanes2021softce}. Combined with the GI distribution described in the previous section, using the volumetric soft cross-entropy loss leads to accuracy gains.

\subsection{Dataset Characteristics} \label{sec:dataset}
One of our model development goals is to deploy models on a camera layout similar to that shown in Fig.~\ref{fig:hardware}. This three-camera plenary setup is the minimum to cover the semi-sphere \fov{} on top of the plane. This setup also ensures that the robot body in the middle of the cameras will not block the view of more than two cameras, making stereo distance estimation possible for all \fov{} directions. Currently, no such dataset exists and it motivates us to create a new dataset. In total, 95K samples were collected from over 60 Unreal Engine 4 simulation environments used in the collection efforts of TartanAir \cite{tartanair2020iros}. This dataset is over 10x larger than any currently available dataset for \omnidirectional{} stereo vision with fisheye images \cite{won2019omnimvs} and is released for download on our project page. The camera layout is the \textit{training layout} in Fig.~\ref{fig:layouts}. Each sample consists of three RGB-dense distance pairs in fisheye format. There are a large variety of outside, urban, indoor, and natural environments as shown in Fig.~\hyperref[fig:datasetbanner]{\ref*{fig:datasetbanner}}.

\section{EXPERIMENTAL PROCEDURE \& RESULTS}

\begingroup
\setlength{\belowcaptionskip}{-10pt}

\begin{figure*}[ht!]
    \centering
    \includegraphics[width=0.95\linewidth]{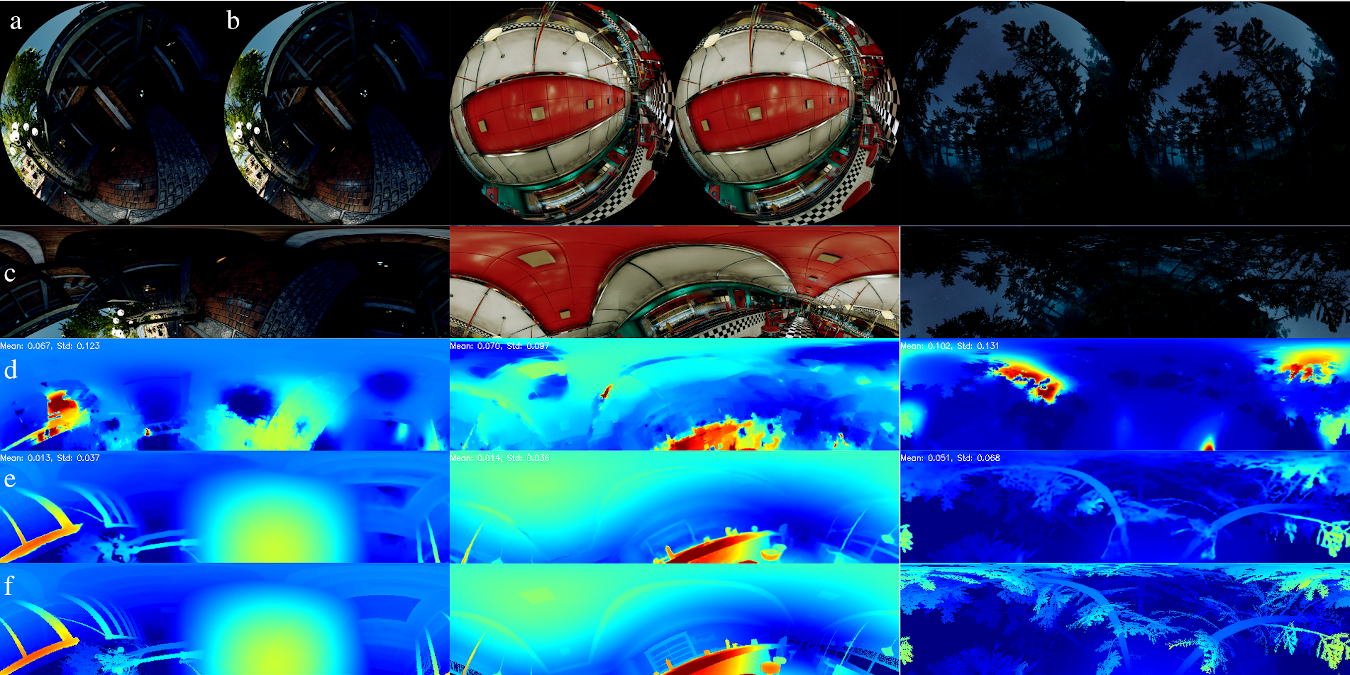}
    \caption{Comparison with synthetically-generated images from the unseen environments. a, b: second and third views. c: equirectangular projected reference (first) view. d, e: outputs of RTSS \cite{meuleman2021real} and G16VV (ours). f: ground truth distance aligned with the reference view. In scenes with low light, high-frequency features such as patterns and trees, and thin objects, G16VV is more accurate and it can resolve fine details.}
    \label{fig:syntheticresults}
\end{figure*}

\endgroup

\subsection{Model Variants and Camera Layouts} \label{sec:model_variants}
We propose that geometry-informed (GI) distance candidates can directly improve distance estimation. GI candidates can be adapted to most of the MVS-\omnidirectional{} vision models where a fixed number of candidates are applied. In this work, we use a set of model variants to show that GI candidates can work well with small candidate numbers and changes in camera layout.

For a baseline comparison, we build a model for the 3-camera layout shown in Fig.~\ref{fig:architecture} using a similar structure as the OmniMVS model\cite{won2019omnimvs}, the state-of-the-art \omnidirectional{} distance estimator. We then have two simple variants from OmniMVS, based on EV and GI candidates. We are targeting models with fewer candidates to have better efficiency. We designate model names E16 and E8 for baseline models with only 16 and 8 candidates, while G16 and G8 for the GI ones. Using the same naming, let G16V be the model trained with the volume loss function. Finally, we also apply the variance cost volume \cite{yao2018mvsnet} to G16V and get G16VV. One detail about G16VV is that when calculating the cost volume, we explicitly handle the self-occlusion from the robot. This is done by additionally showing the model a binary mask for every input fisheye image. Such a mask marks non-occluded pixels as valid pixels. When building the cost volume, a variance value is calculated by only considering visual features from the non-masked regions. G16VV is smaller than other variants as a result of using feature variance for building the cost volume. Besides E16 and E8, we also make compare with the RTSS model \cite{meuleman2021real}. In addition to the original RTSS model with 32 EV candidates, we also tested RTSS with GI candidates and 16-candidate variants. 

All models are trained on the dataset in Section.~\ref{sec:dataset}. The distance range is fixed at 0.5-100m during training. For comparison purposes, all models are trained with the same fixed learning rate (0.0001) and batch size (16). We reserve some simulation environments from training and collect ground truth data for evaluation. 

\begin{figure}[!b]
    \centering
    \includegraphics[width=0.95\linewidth]{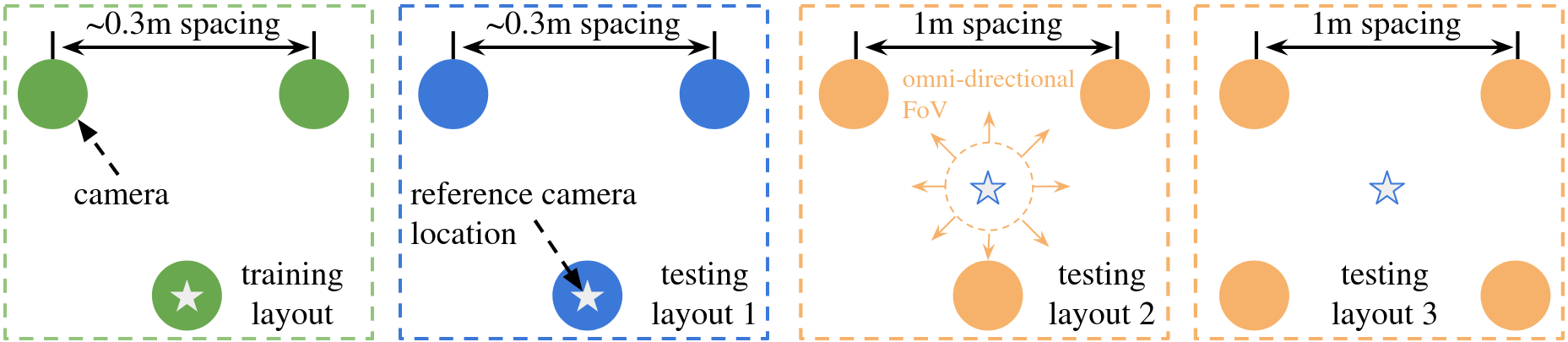}
    \caption{Camera layouts in experiments. From left to right: training, testing 1 (same as training), testing 2 (larger spacing, new reference location), testing 3 (larger spacing, new reference location, and more cameras).}
    \label{fig:layouts}
\end{figure}

Several camera layouts are used in the following experiments. As shown in Fig.~\ref{fig:layouts}, all models are trained using the \textit{training layout}. We test the models on different layouts representing the change of spacing, number of cameras, and reference location. A location on the plane is picked as the reference and the true \omnidirectional{} distance image is generated w.r.t this reference location in the simulator.

\subsection{Evaluation with the Same Camera Layout}

We first collected over 1000 samples using \textit{testing layout 1} in Fig.~\ref{fig:layouts}. Model predictions are compared with ground truth \omnidirectional{} distance images. We use simple metrics including mean absolute error (MAE), mean root square error (RMSE), and the Structural Similarity Index (SSIM) as in \cite{meuleman2021real}. All metrics are computed using the inverse distance (ranging from 0.01 to 2). We use a single NVIDIA V100 GPU for measuring the execution time and GPU memory usage. Several observations can be made from Table~\ref{tab:same_layout}: 1) with 16 or 8 candidates, a model can have very competitive efficiency and GPU consumption compared to the real-time baseline model (RTSS\cite{meuleman2021real}, model RS-E16 to RS-G32). 2) GI candidates do not improve the non-learning baseline model\cite{meuleman2021real} and this is the expected behavior. The baseline's performance increases with more candidates. 3) When using very few candidates, such as E8 and G8, the one that uses GI candidates tends to be better. 4) Upon proper training, all learning-based models have similar performance with and without GI candidates, if tested using the same camera layout. Until now, GI candidates have shown marginal performance gain. However, significant improvement is shown in the next section where the testing camera layout is different from the training configuration.

% The following table has been updated by Yaoyu using the latest config21 result. Now G16V is config21 (instead of config29). 
% The text of the paper is now consistent with the table.
\begin{table}[t]
\scriptsize
\caption{Comparison using the same camera layout}
\begin{tabular}{ccc|ccc|c|cc}
\hline
\multirow{2}{*}{model} & \multicolumn{2}{c|}{candidates} & \multicolumn{3}{c|}{metrics} & time                     & \multicolumn{2}{c}{GPU (MB)}                  \\ \cline{2-9} 
                       & type            & num           & MAE      & RMSE    & SSIM    & (ms)                     & start                 & peak                  \\ \hline
RS-E16                 & EV              & 16            & 0.075    & 0.129   & 0.699   & 146                      & \multirow{2}{*}{820}  & \multirow{2}{*}{2780} \\
RS-G16                 & GI              & 16            & 0.076    & 0.129   & 0.713   & 140                      &                       &                       \\ \hline
RS-E32                 & EV              & 32            & 0.053    & 0.101   & 0.776   & 144                      & \multirow{2}{*}{1250} & \multirow{2}{*}{5130} \\
RS-G32                 & GI              & 32            & 0.059    & 0.105   & 0.777   & \multicolumn{1}{l|}{146} &                       &                       \\ \hline
E8                     & EV              & 8             & 0.013    & 0.032   & 0.862   & \multirow{2}{*}{65}      & \multirow{2}{*}{790}  & \multirow{2}{*}{1030} \\
G8                     & GI              & 8             & 0.012    & 0.029   & 0.867   &                          &                       &                       \\ \hline
E16                    & EV              & 16            & 0.011    & 0.028   & 0.876   & \multirow{3}{*}{111}     & \multirow{3}{*}{790}  & \multirow{3}{*}{1230} \\
G16                    & GI              & 16            & 0.010    & 0.028   & 0.877   &                          &                       &                       \\
G16V                   & GI              & 16            & 0.013    & 0.028   & 0.875   &                          &                       &                       \\ \hline
G16VV                  & GI              & 16            & 0.012    & 0.028   & 0.872   & 114                      & 800                   & 1090                  \\ \hline
\end{tabular}
\\ \textit{EV}: evenly distributed candidates. \textit{GI}: geometry-informed. \textit{RS}: the RTSS\cite{meuleman2021real} model.
\label{tab:same_layout}
\end{table}

%\begin{figure}[!h]
%    \centering
%    \includegraphics[width=0.9\linewidth]{figures/main_eval.png}
%    \caption{Sample of omni-directional distance output. From top to bottom: prediction, ground truth, three equi-rectangular warped input fisheye images. Numbers at top-left: mean and standard deviation of absolute error. \cmtyh{put the fisheye images to the left, and draw camera layout?}}
%    \label{fig:datasetbanner}
%\end{figure}

\begingroup
\setlength{\belowcaptionskip}{-10pt}

\begin{figure*}[!t]
    \centering
    \includegraphics[width=0.95\linewidth]{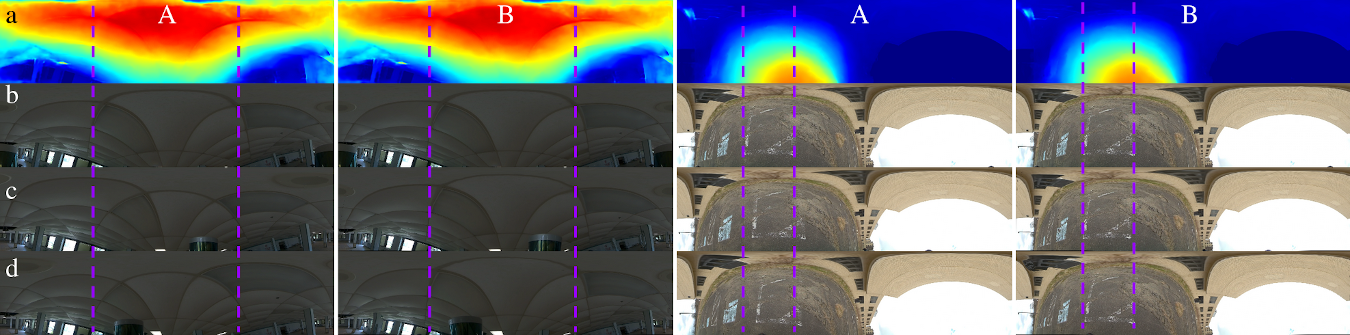}
    \caption{Sample results on real-world data. Model G16VV. Row a-d: distance estimation, three camera views. Columns: A - Original input image, B - Input image warped using the predicted distance (Raw a). Purple lines: vertical guidelines. If the distance prediction is good, then the pixels on the purple line across Column B should align among Row b-d. Our model trained only on synthetic data evaluated on real images. The model can be optimized with NVIDIA TensorRT for better inference speed on real robots. }
    \label{fig:realdata}
\end{figure*}

\endgroup

\subsection{Evaluation with Different Camera Layout}

\begin{figure}[!b]
   \centering
   \includegraphics[width=0.9\linewidth]{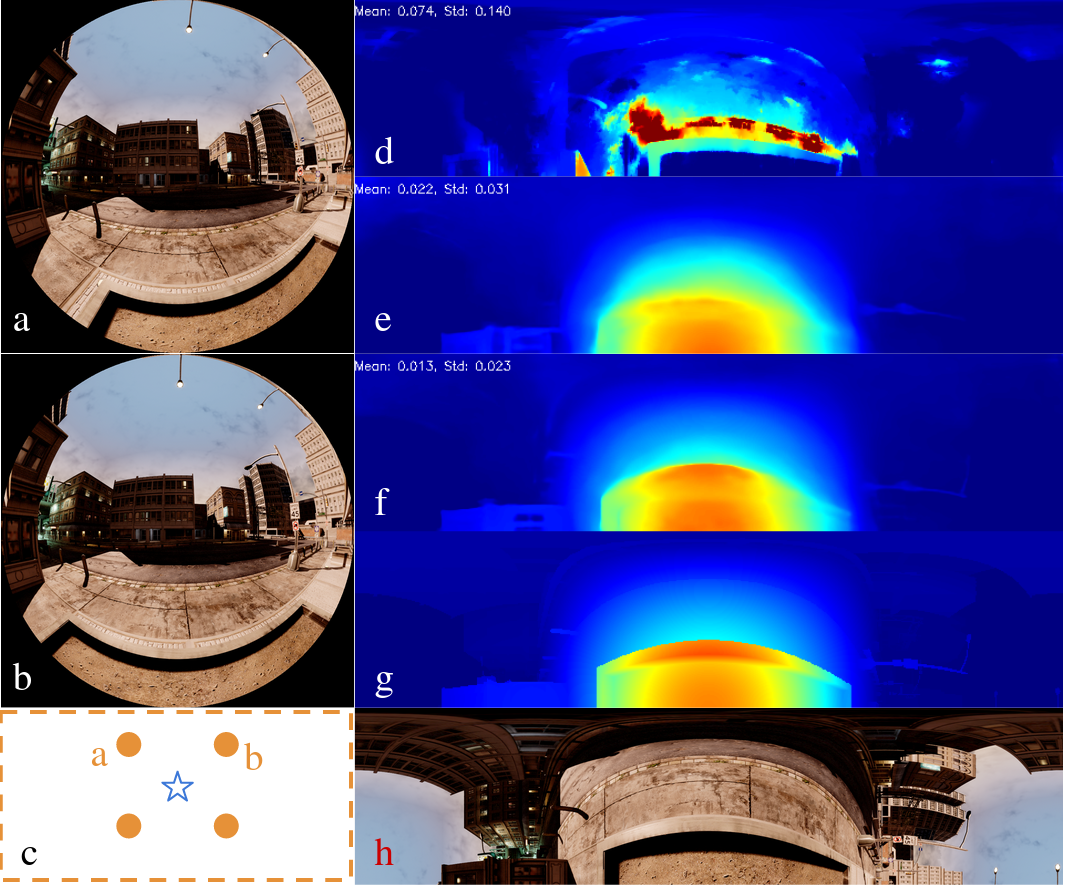}
   \caption{Sample results of four-camera layout. a: first view. b: second view. c: camera layout. d: RTSS, e: G16VV w/ training candidates. f: G16VV w/ adjusted GI candidates calculated w.r.t. camera layout. g: ground truth distance map. h: ground truth equirectangular image view. When adjusted GI candidates are used, G16VV is more accurate and resolves more details.}
   \label{fig:4_cam}
\end{figure}

To show the GI candidates' ability to handle a camera layout that is different from the training setup, we collect over 100 samples from the evaluation environments with larger distances among the cameras, as illustrated in \textit{testing layout 2} Fig.~\ref{fig:layouts}. For this test, we only use the variants that have 16 candidates. In the tests, we apply a trained model twice, one with the candidates it was trained on, and the other with the dedicated new candidates that are calculated concerning the deployed camera layout (denoted as \textit{new} in the following table). Table~\ref{tab:diff_layout} shows that the GI candidates can boost the performance of a trained model when deployed on a camera layout that has longer displacement than the training data. We also observe from Table~\ref{tab:diff_layout} that model G16V, which is trained with our volume loss, tends to have better SSIM values. 

Using the G16VV model, since it builds the cost volume with feature variance across all views, we can demonstrate that using the GI candidates, our model can also handle the change of camera number. A separate set of over 100 samples is collected from evaluation environments with four cameras laid out as \textit{testing layout 3} in Fig.~\ref{fig:layouts}. We show a sample result in Fig.~\ref{fig:4_cam}. The quantitative results are also listed in Table~\ref{tab:diff_layout}. G16VV gains better performance from only changing the candidate values without any new training. On the speed side, from 3 cameras to 4 cameras, the processing time of G16VV adds only 6ms while the RTSS model experiences a 35ms time increase. On the GPU memory side, since the RTSS model precomputes the best view pairs for every output pixel, its memory does not change. For G16VV, we observe an increase of about 60M Bytes.

% The following table has been updated by Yaoyu using the latest config21 result. Now G16V is config21 (instead of config29). 
% The text of the paper is now consistent with the table.
\begin{table}[h]
\centering
\caption{Comparison using new camera layouts}
\label{tab:diff_layout}
\begin{tabular}{ccccccc}
\hline
\multirow{2}{*}{\begin{tabular}[c]{@{}c@{}}new \\ layout\end{tabular}}     & \multirow{2}{*}{model} & \multicolumn{2}{c}{candidates} & \multicolumn{3}{c}{metrics} \\ \cline{3-7} 
                                                                           &                        & train          & eval          & MAE     & RMSE    & SSIM    \\ \hline
\multirow{7}{*}{\begin{tabular}[c]{@{}c@{}}3 cam \\ 1m apart\end{tabular}} & RS-E32                 & -              & EV            & 0.124   & 0.217   & 0.697   \\ \cline{2-7} 
                                                                           & \multirow{2}{*}{E16}   & EV             & EV            & 0.033   & 0.053   & 0.768   \\
                                                                           &                        & EV             & new           & 0.018   & 0.037   & 0.829   \\ \cline{2-7} 
                                                                           & \multirow{2}{*}{G16}   & GI             & GI            & 0.030   & 0.050   & 0.786   \\
                                                                           &                        & GI             & new           & 0.020   & 0.039   & 0.823   \\ \cline{2-7} 
                                                                           & \multirow{2}{*}{G16V}  & GI             & GI            & 0.030   & 0.048   & 0.783   \\
                                                                           &                        & GI             & new           & 0.020   & 0.038   & 0.837   \\ \hline
\multirow{3}{*}{\begin{tabular}[c]{@{}c@{}}4 cam \\ 1m apart\end{tabular}} & RS-E32                 & -              & EV            & 0.090   & 0.147   & 0.637   \\ \cline{2-7} 
                                                                           & \multirow{2}{*}{G16VV} & GI             & GI            & 0.024   & 0.041   & 0.817   \\
                                                                           &                        & GI             & new           & 0.016   & 0.033   & 0.860   \\ \hline
\end{tabular}
\\ Candidates type: \textit{EV} - evenly distributed, \textit{GI} - geometry informed, \textit{new} - GI for the 1m spacing. All models are trained with a camera spacing of about 0.3m and tested with 1m. \textit{RS}: the RTSS\cite{meuleman2021real} model.
\end{table}

\subsection{Deployment \& Hardware Acceleration}

We deploy the G16VV model on several Nvidia Jetson devices. To leverage the hardware acceleration capability, we convert the entire model using TensorRT. We demonstrate that our optimized model is capable of achieving more than \SI{10}{\hertz} on AGX Orin as shown in Table~\ref{tab:deploy}. All the deployed models (and intermediate, hardware-independent models) are available on the project website.

\begin{table}[h]
\centering
\caption{Inference memory and time of accelerated models.}
\label{tab:deploy}
\begin{tabular}{cccc}
\hline
\multirow{2}{*}{architecture}  & \multirow{2}{*}{GPU} & \multicolumn{2}{c}{inference} \\ \cline{3-4} 
                               &                      & time (ms)     & mem. (MB)     \\ \hline
\multirow{2}{*}{x86-64}        & RTX3080Ti            & 11            & 710           \\
                               & GTX1070MQ            & 210           & 800           \\ \hline
\multirow{3}{*}{NVIDIA Jetson} & AGX Xavier           & 200           & 600           \\
                               & Xavier NX            & 270           & 1800          \\
                               & AGX Orin             & 65            & 1900          \\ \hline
\end{tabular}

Original model: G16VV. The inference time and memory consumption are measured across 100 consecutive runs. The entire model is accelerated by TensorRT and all the models are available from the project website.

\end{table}

\section{CONCLUSION}

This work introduces Geometry-Informed (GI) distance candidate selection for \omnidirectional{} vision models. GI candidate approximate constant feature displacement between distance candidates. Additionally, GI candidates give the model extra flexibility after training: camera spacings (stereo baselines) can be adjusted after training without fine-tuning while maintaining good performance. We develop a set of models with our improvements and compare them against available state-of-the-art baseline models and show accuracy, speed, and memory consumption improvements. 
% We also show that our model, only trained on synthetic data, produces reasonable distance estimations in the real world. 
Lastly, we release several model variants and our dataset for the use by the community.

                                  % the text height of the last page by a suitable amount.

%%%%%%%%%%%%%%%%%%%%%%%%%%%%%%%%%%%%%%%%%%%%%%%%%%%%%%%%%%%%%%%%%%%%%%%%%%%%%%%%

\section*{ACKNOWLEDGMENT}

Special thanks to Wenshan Wang and the TartanAir team for providing the latest simulation environments. Special thanks to Siheng Teng regarding hardware deployment. This work was funded by the Defence Science and Technology Agency, Singapore. This work used Bridges-2 at PSC through allocation cis220039p from the Advanced Cyberinfrastructure Coordination Ecosystem: Services \& Support (ACCESS) program which is supported by NSF grants \#2138259, \#2138286, \#2138307, \#2137603, and \#213296.

\bibliography{citations}
\bibliographystyle{ieeetr}

\end{document}